\documentclass{article}
\usepackage[dvipsnames]{xcolor}
\usepackage{svg}

\usepackage{placeins}
\usepackage{algpseudocode,algorithm,algorithmicx}
\usepackage[toc,page]{appendix}
\usepackage{hyperref}
\usepackage{framed}
\usepackage{flafter}
\usepackage{amsthm}
\usepackage{geometry}

\newcommand*\mathsub[1]{_{\textrm{\tiny #1\normalsize}}}
\algdef{SE}[DOWHILE]{Do}{doWhile}{\algorithmicdo}[1]{\algorithmicwhile\ #1}%

\newcommand{\numsuccessfullcuts}{304 }

\newcommand{\numcubicles}{144 }

\newcommand{\exptwidth}{0.25}

\newcommand{\widthmultiplierdraw}{0.15 }
\newcommand*\Let[2]{\State #1 $\gets$ #2}

\newcommand*\localoptstart{\color{blue}}
\newcommand*\localoptend{\color{black}}

\newcommand*\focusstart{\color{ForestGreen}}
\newcommand*\focusend{\color{black}}

\algrenewcomment[1]{\focusstart \hspace{1em} \(//\) #1 \focusend}
\algnewcommand{\LineComment}[1]{\State \(//\) #1}

\title{Integrating asymptotically-optimal path planning with local optimization}
\author{Scott Paulin, Tom Botterill, XiaoQi Chen, Richard Green}
\begin{document}
	\maketitle
\begin{abstract}
%In order for robots to be effective, they need to be able to quickly compute high quality motion plans. 

Many robots operating in unpredictable environments require an online path planning algorithm that can quickly compute high quality paths. Asymptotically optimal planners are capable of finding the optimal path, but can be slow to converge. Local optimisation algorithms are capable of quickly improving a solution, but are not guaranteed to converge to the optimal solution. In this paper we develop a new way to integrate an asymptotically optimal planners with a local optimiser. We test our approach using RRTConnect* with a short-cutting local optimiser. Our approach results in a significant performance improvement when compared with the state-of-the-art RRTConnect* asymptotically optimal planner and computes paths that are 31\% faster to execute when both are given 3 seconds of planning time.
\end{abstract}

\section{Introduction}
A path planning algorithm finds a collision free path for a robot to follow in order to perform a task, for example to move a robot arm. Robots that perform tasks in uncontrolled environments, e.g. autonomous driving~\cite{Thrun2006} or agricultural tasks~\cite{Bac2014,Lee2014a,Scarfe2009}, must plan new paths online for each task, as they identify goals and obstacles to avoid. These path planners must be both computationally efficient, so they can plan paths with a limited time budget, and must find short fast-to-execute paths. Ideally, the planner should find paths that are as close to the shortest/optimal path that is possible.

A popular family of algorithms for planning paths for robot arms are randomized sampling-based path planners, e.g. Probabilistic Road Maps (PRMs)~\cite{Kavraki1996} and Rapidly-exploring Random Trees (RRTs)~\cite{LaValle1998}. Many of algorithms find feasible paths quickly but are not guaranteed to find the shortest path, regardless of time available. Recent work on optimal planning, e.g. RRT*\cite{Karaman2011} and PRM*~\cite{Karaman2011}, extend these algorithms to guarantee asymptotic optimality, however, these algorithms may require a long time to find a good path. This paper considers speeding-up existing optimal planning algorithms for practical applications where the computation time available for planning is limited.

This paper evaluates optimal path planners for the problem of using a six degree-of-freedom robot arm to reach and prune (cut) a grape vine, and the problem of reaching into cubicles. We propose improving convergence speed by integrating a local `short-cutting' optimiser to improve intermediate solutions. For these applications we demonstrate that combining RRTConnect*~\cite{Akgun2011, Jordan2013, Klemm2015} (a bidirectional variation of RRT*) and short-cutting results in substantially faster convergence. 

%Maybe RRTConnect*WS
\begin{table}[htbp]
	\centering
	\caption{Summary of planners evaluated}
	\begin{tabular}{|p{3cm}|p{7cm}|}
		\hline
		\textbf{Name} & \textbf{Description} \\ \hline
		RRTConnect+S & One run of RRTConnect followed by short-cutting optimizer\\ \hline
		MRRTConnect+S & Multiple restarts of RRTConnect + short-cut where the best path is kept. A leading contemporary approach~\cite{Luo2014}. See Fig.~\ref{alg:m_rrt_connect_short}.\\ \hline
		RRTConnect* & Asymptotically optimal RRTConnect. We also use the informed heuristics~\cite{Gammell2014}.\\ \hline
		RRTConnect*+S & Our proposed approach. RRTConnect* with informed heuristics that uses short-cutting local optimizer.\\ \hline
	\end{tabular}
	\label{tab:planners}
\end{table}

\section{Background}
Robot arm path planners often operate in the robot's \emph{configuration space}~\cite{Lozano-perez1983} to find collision free paths. The configuration space, $C$, can be split into $C\mathsub{free}$ and $C\mathsub{obs}$. $C\mathsub{free}$ is the set of all configurations where the robot is not in collision with the environment or itself. $C\mathsub{obs}$ is the set of configurations where the robot is in collision with itself or the environment. Computing an explicit representation of configuration space is prohibitively expensive for many robot arms.

Sampling based motion planners~\cite{Kavraki1996, LaValle1998, Sucan2010a, Sucan2012} have become popular because they do not require an explicit representation of the robot's configuration space. These planners use a collision detector to classify sampled configurations as either in $C\mathsub{free}$ or $C\mathsub{obs}$. These planners explore the robot's configuration space and grow a graph. Some planners, e.g. PRM, construct a highly connected graph that can be used for multiple planning queries. Other planners, e.g. RRT, quickly grow a directed acyclic graph that can only be used for one planning query.

Path planners can be categorised as feasible planners~\cite{Hsu1997,Kavraki1996,LaValle1998,LaValle2001}, or optimizing planners~\cite{Gammell2015,Janson2015,Karaman2011, Salzman2016}. Feasible planners attempt to quickly find a solution and terminate as soon as one is found. Feasible planners can return poor, e.g. long, solutions because they do not perform optimization. Optimizing planners attempt to find high-quality, e.g. short, solutions within a set computation time or number of iterations. Some of these optimizing planners are asymptotically optimal and will converge to the optimal solution eventually~\cite{Gammell2015,Janson2015,Karaman2011}. Other optimizing planners are asymptotically near-optimal and will converge to a near-optimal solution eventually~\cite{Dobson2014, Salzman2016}. In this paper we propose an approach for speeding up the convergence of optimizing planners.

A key requirement for many popular asymptotically optimal path planners, e.g. Batch Informed Trees (BIT*)~\cite{Gammell2015}, RRT*, PRM*, is that when a new vertex $v$ is inserted into the planner's graph $G$, edges are formed between $v$ and vertices within its neighbourhood $V\mathsub{nbh}$. $V\mathsub{nbh}$ can be defined as all vertices in $G$ within a radius $r\mathsub{nbh}$ of $v$, or as the $k\mathsub{nbh}$ neighbours of $v$. Minimum values for $r\mathsub{nbh}$ and $k\mathsub{nbh}$ depend on the number of vertices in $G$~\cite{Karaman2011}. Some multiple query planners, e.g. PRM*, form edges between $v$ and all its neighbours where a collision-free path exists. Single query planners perform a \emph{rewiring} step that joins $v$ to a sub-set of it's neighbours without adding cycles their graph.

\begin{figure}[htbp]
	\centering
	\newgeometry{textwidth=20cm}
	\hspace*{-1.75in}
	\subfloat[]{\fbox{\includesvg[ width=\widthmultiplierdraw\linewidth]{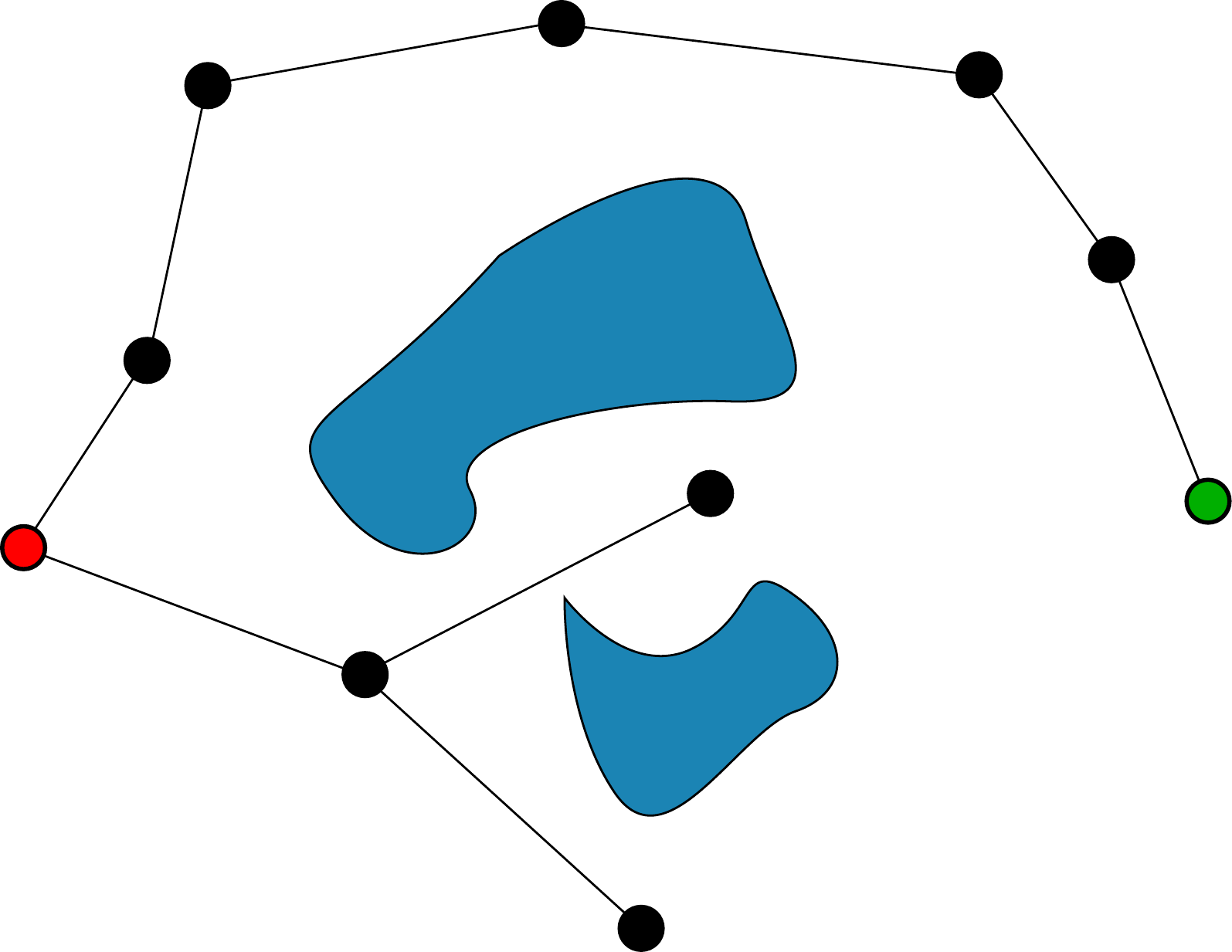}}\label{fig:rrtstar_rewire_G}}\hspace{1em}
	\subfloat[]{\fbox{\includesvg[ width=\widthmultiplierdraw\linewidth]{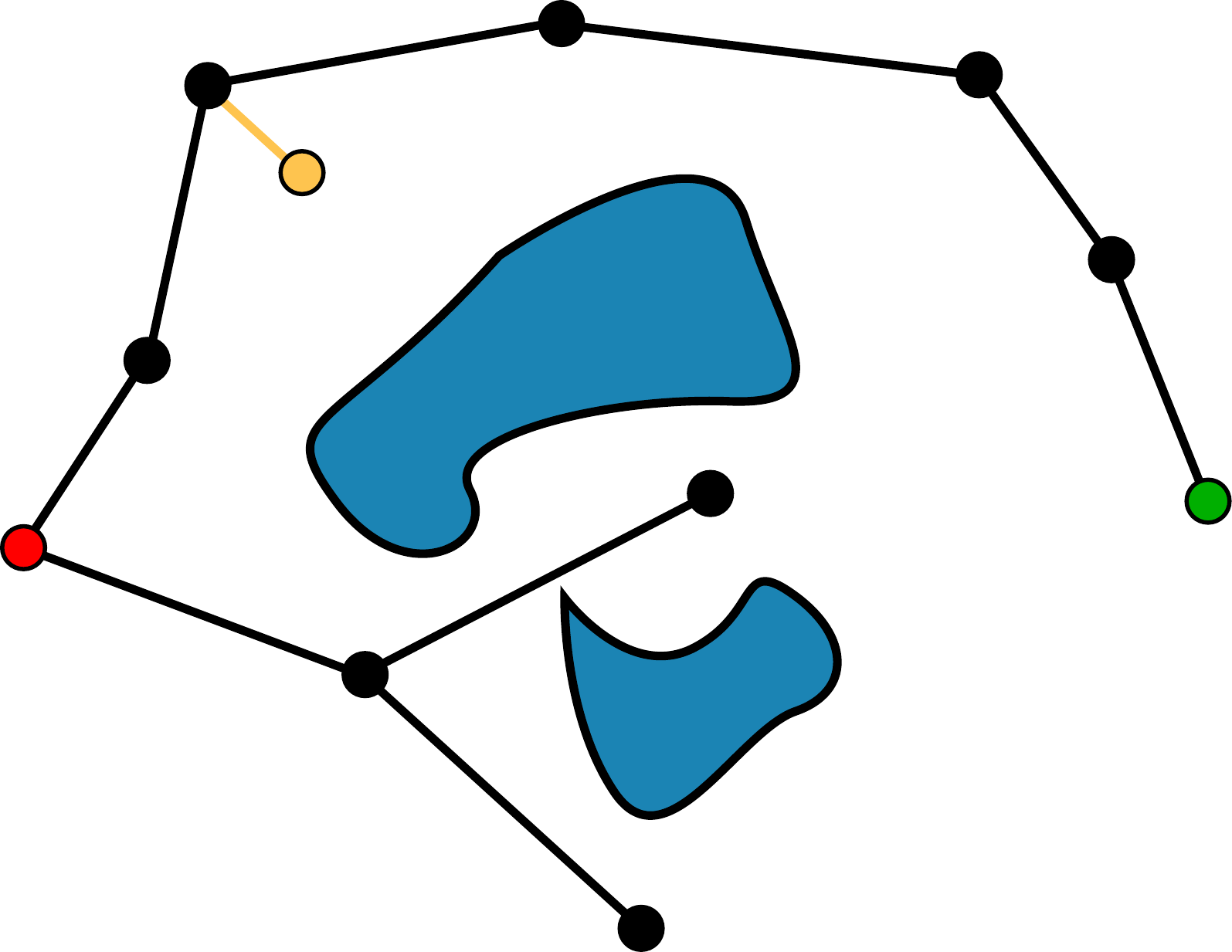}}\label{fig:rrtstar_rewire_add_v}}\hspace{1em}
	\subfloat[]{\fbox{\includesvg[ width=\widthmultiplierdraw\linewidth]{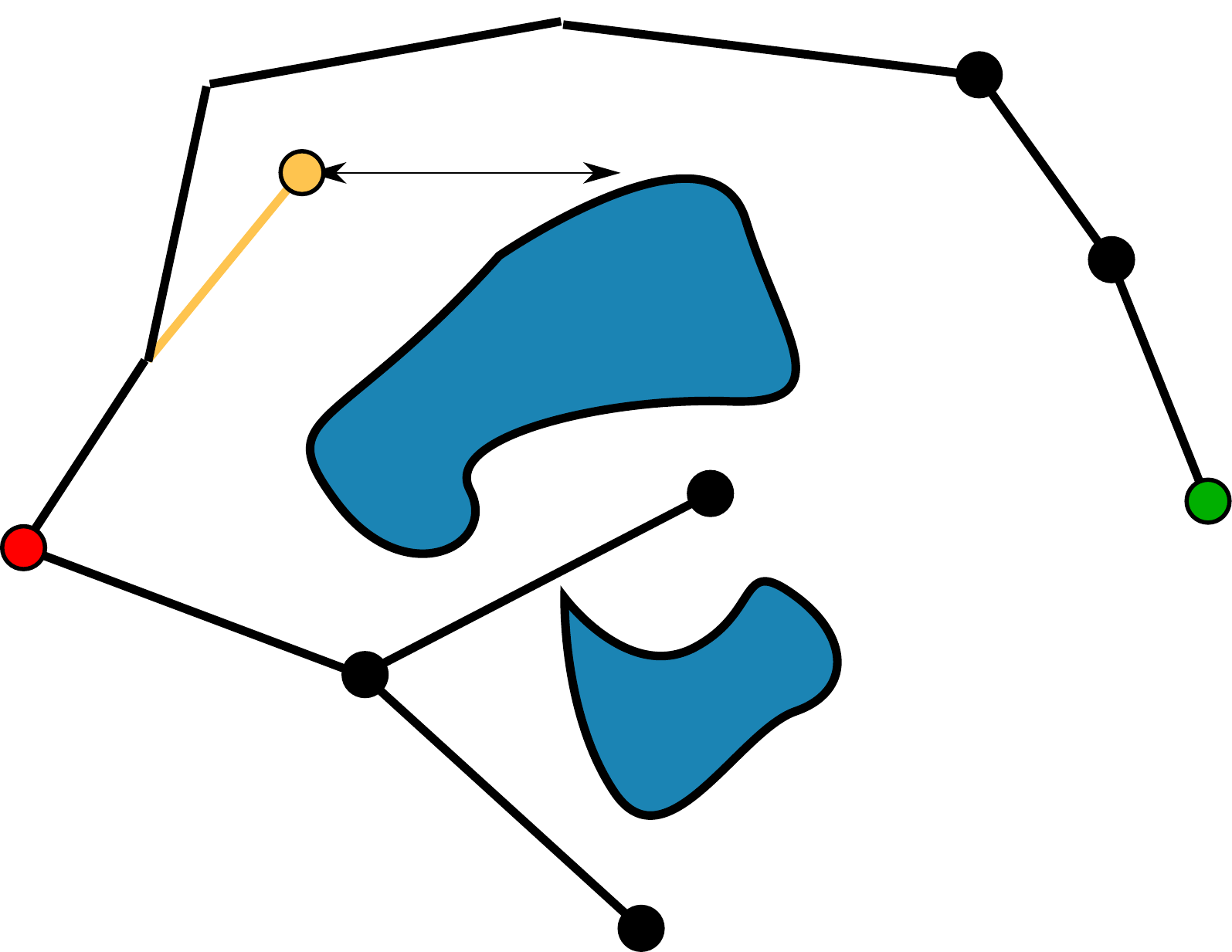}}\label{fig:rrtstar_rewire_sbp}}\hspace{1em}
	\subfloat[]{\fbox{\includesvg[ width=\widthmultiplierdraw\linewidth]{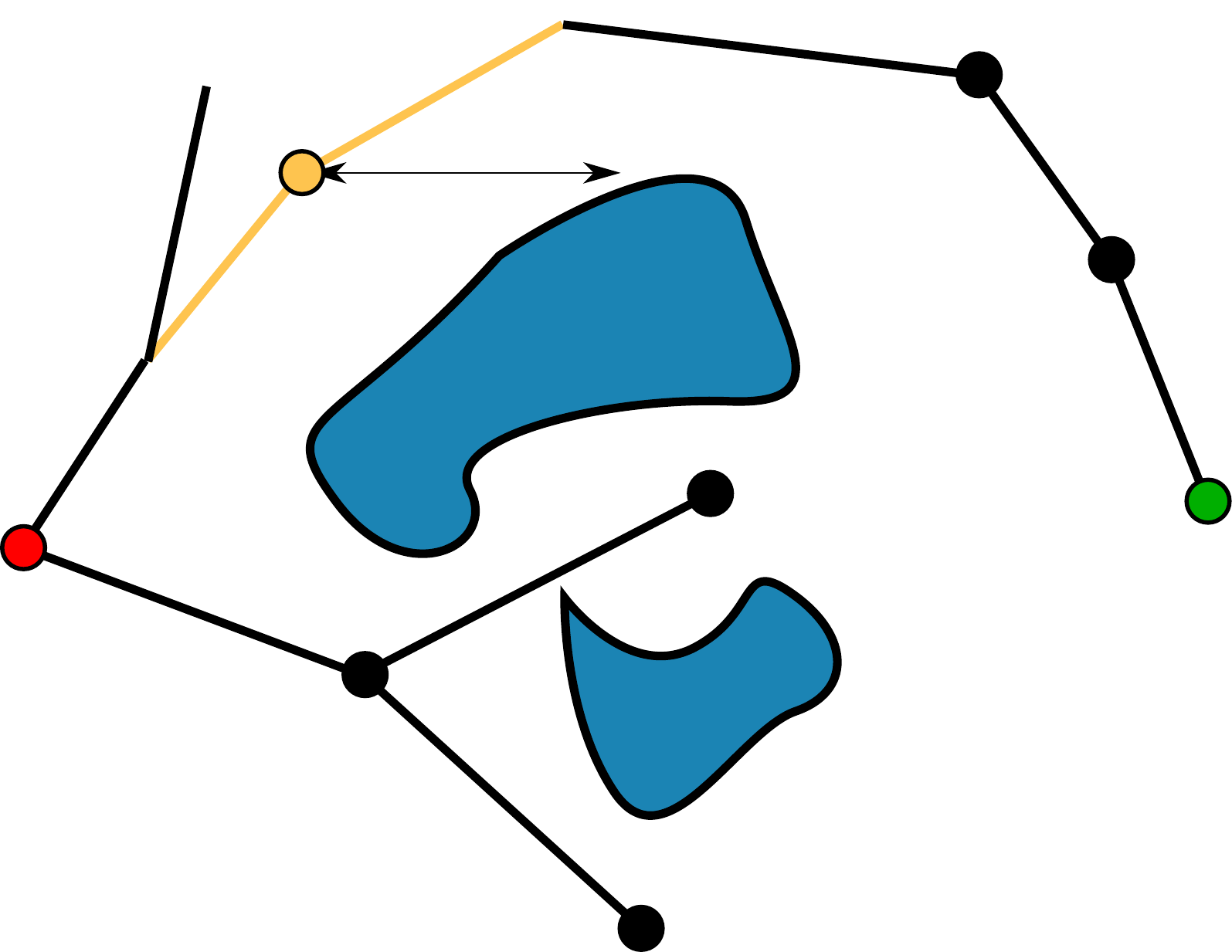}}\label{fig:rrtstar_rewire_update_nbh}}
	\restoregeometry
	\caption{RRT* insertion of the yellow vertex. The start vertex is shown in red and the goal vertex is shown in green. $r_n$ denotes the neighbourhood radius and vertices within this neighbourhood are shown with a white fill.}
	\label{fig:rrtstar_rewire}
\end{figure}

Fig.~\ref{fig:rrtstar_rewire} illustrates how RRT* rewires a new vertex $v$ into its graph $G$ (Fig.~\ref{fig:rrtstar_rewire_G}). The new vertex is initially connected to its nearest neighbour in $G$ (Fig.~\ref{fig:rrtstar_rewire_add_v}). The neighbourhood of $v$ in $G$ is computed, $v$ is then connected to a different vertex to minimise the cost to arrive from the start vertex (Fig.~\ref{fig:rrtstar_rewire_sbp}). The edges of neighbouring vertices are changed if the path through $v$ provides smaller cost to arrive at that particular neighbour (Fig.~\ref{fig:rrtstar_rewire_update_nbh}). This rewiring approach is guaranteed not to introduce cycles into $G$~\cite{Karaman2011}.

As optimising planners converge to the optimal solution they can spend more time processing samples that cannot possibly be used to improve on the best solution~\cite{Gammell2014}. This can be remedied by focussing the planner's search to useful regions of configuration space~\cite{Gammell2014} and rejecting new samples that cannot be used to improve on the planner's best solution~\cite{Akgun2011} without sacrificing the optimality properties of the planner. In this paper we test RRTConnect* with these heuristics enabled.

Local optimization algorithms can be used with path planners to quickly improve path quality, e.g. length. These algorithms often rely on a path planner to provide an initial solution. This initial solution influences the quality of the optimised path because these algorithms only optimize locally. Short-cutting~\cite{Berchtold1994,Geraerts2007,Hauser2010} for reducing path length and sequential convex optimization approaches~\cite{Kalakrishnan2011, Schulman2014, Zucker2013} have been shown to work well on robot arms.

A common approach to finding short paths is to find an initial collision-free solution with a feasible planner, e.g. RRTConnect~\cite{Kuffner2000} (a bidirectional RRT), and to optimise this path with a local optimiser e.g. short-cutting. Another approach is to perform multiple restarts of the feasible planner, optimise each solution and return the best solution as shown in Fig.~\ref{alg:m_rrt_connect_short}. This has been shown to work well in empirical experiments when compared to asymptotically optimal planners~\cite{Luo2014}. We compare our approach of RRTConnect* integrated with short-cutting to multiple restarts of RRTConnect with short-cutting, as well as one run of RRTConnect with short-cutting.

\begin{figure}[htbp]
	\begin{framed}
		\begin{algorithmic}[1]%TODO mention implementation diffs to empiricle paper.
			\Function{MRRTConnect+S}{$v\mathsub{start},V\mathsub{goal}$, termination\textunderscore condition}
			\Let{$L\mathsub{best}$}{$\infty$}
			\Let{$p\mathsub{best}$}{NULL}
			\Do
			    \Let{$p$}{RRTConnect($v\mathsub{start},V\mathsub{goal}$)}
			    \Let{$p$}{Shortcut($p$)}
			    \Let{$L$}{The length of $p$}
			    \If{$L < L\mathsub{best}$}
			        \Let{$L\mathsub{best}$}{$L$}
			        \Let{$p\mathsub{best}$}{$p$}
			    \EndIf
			\doWhile{not termination\textunderscore condition}
			\State \Return $p\mathsub{best}$
			\EndFunction
		\end{algorithmic}
	\end{framed}
	\caption{Multiple restarts of RRTConnect with short-cutting.}\label{alg:m_rrt_connect_short}
\end{figure}

There has been some recent interest in combining asymptotically optimal path planners with local optimisers to speed up convergence to optimal solutions. Choudhury et. al.~\cite{Choudhury2016} use the CHOMP local optimiser to avoid collisions in edges between vertices in their Regionally Accelerated Batch Informed Trees (RABIT*) planner. This differs from our work because our approach uses a local optimiser to improve a complete path. 

A recent preprint proposed `interleaving' the use of a global asymptotically optimal path planner with a local optimiser~\cite{Kuntz2016}. The global planner is used to explore the robot's configuration space and the local optimiser is used to quickly improve solutions. The local optimiser is invoked every time the global planner finds a better solution. The optimised path is then placed into the planner's graph without forming edges to existing vertices within the graph i.e. no rewiring step is performed. This means that the interleaving approach is only asymptotically optimal for some planners, e.g. PRM*, and special consideration must be given to only include vertices added by the global planner when calculating the neighbourhood size.

In this paper we build on the interleaving approach in two ways: Firstly, optimised paths are rewired back into the planner's graph to preserve the asymptotic optimality of the global planner. Secondly, we only invoke the local optimiser when the global planner has substantially improved on the last optimised path. This prevents the local optimiser being invoked every time the global planner has made a small incremental improvement to the last optimised path.

\section{Integrating RRTConnect* with a short-cutting local optimiser}
\label{sec:integration}
Our approach speeds up RRTConnect* by integrating a short-cutting local optimiser. Good intermediate solutions found by RRTConnect* are shortcut and inserted into RRTConnect*'s graph as shown in Fig.~\ref{alg:integrated_with_local_opt}. $v\mathsub{start}$ and $V\mathsub{goal}$ represent the start vertex and goal vertices for the planning query. Planning continues until the termination\textunderscore condition expires, e.g. this could be an iteration count or a timeout.

\begin{figure}[htb]
	\begin{framed}
		\begin{algorithmic}[1]
			\Function{RRTConnect*+S}{$v\mathsub{start},V\mathsub{goal}$, opt\textunderscore threshold, termination\textunderscore condition}
			\Let{$C\mathsub{last\mathunderscore opt}$}{$\infty$}
			\Do
			\Let{G}{RRTConnect*($v\mathsub{start},V\mathsub{goal}$, G)}\Comment{One iteration}
			\Let{best\textunderscore path}{Best cost path from $v\mathsub{start}$ to $V\mathsub{goal}$ through $G$}
			\Let{$C\mathsub{best}$}{Cost of best\textunderscore path}
			\localoptstart
			\If{$\frac{C\mathsub{last\mathunderscore opt}-C\mathsub{best}}{C\mathsub{last\mathunderscore opt}} >$ opt\textunderscore threshold}
			\Let{$p\mathsub{optimized}$}{Shortcut($p\mathsub{shortest}$)}
			\Let{$G$}{InsertPath($G, p\mathsub{optimised}, v\mathsub{start}$)}
			\Let{$C\mathsub{last\mathunderscore optimized}$}{$C\mathsub{shortest}$}
			\EndIf
			\localoptend
			\doWhile{not termination\textunderscore condition}
			\State \Return Lowest cost path from $v\mathsub{start}$ to $V\mathsub{goal}$ through $G$
			\EndFunction
		\end{algorithmic}
	\end{framed}
	\caption{RRTConnect* with short-cutting. The blue lines show our proposed changes to RRTConnect*.}\label{alg:integrated_with_local_opt}
\end{figure}

\begin{figure}[H]
	\centering
	\newgeometry{textwidth=20cm}
	\hspace*{-1.75in}
	\subfloat[Planner's existing graph $G$ with one solution.]{\fbox{\includesvg[ width=\widthmultiplierdraw\linewidth]{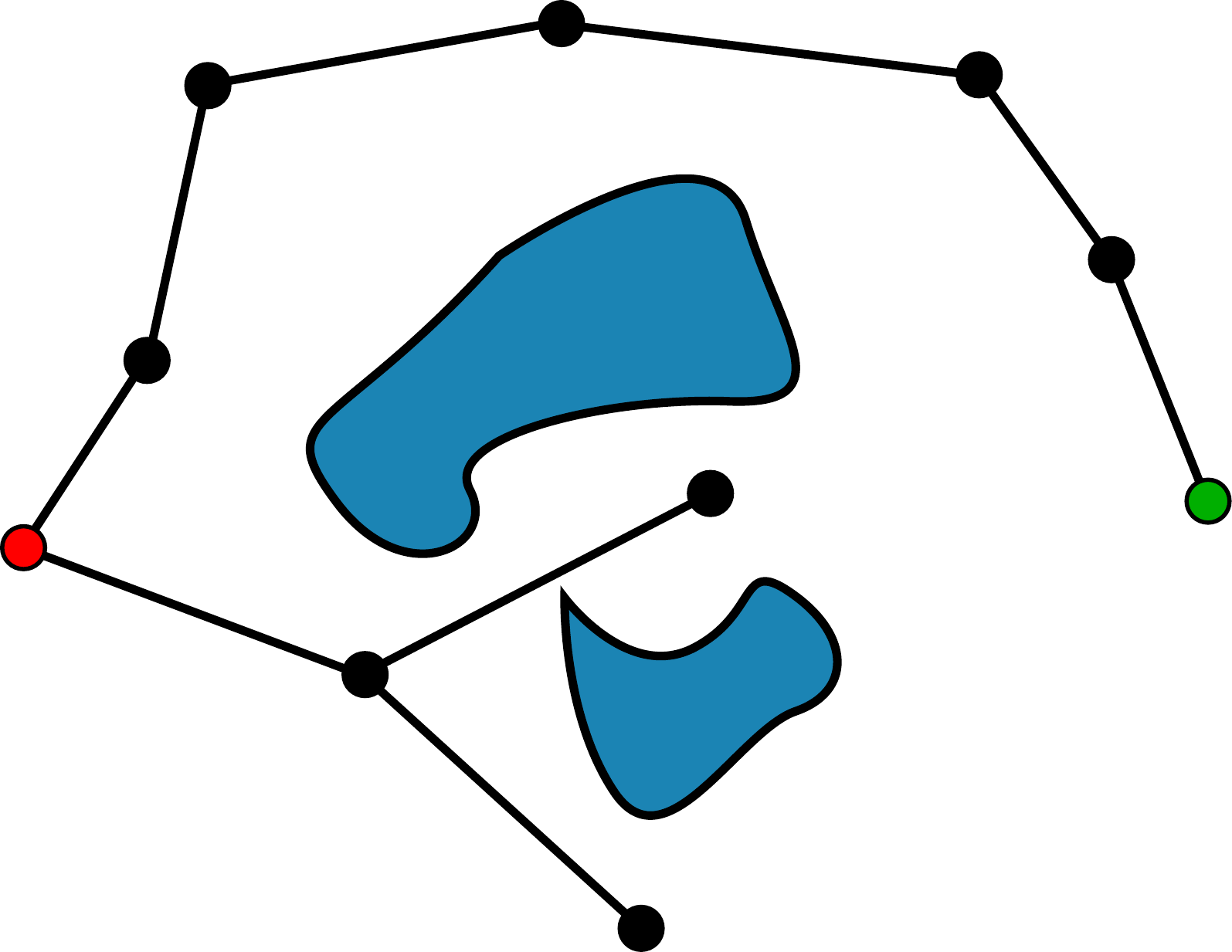}}\label{fig:insert_G}}\hspace{1em}
	\subfloat[Path to be inserted, $p$, into $G$.]{\fbox{\includesvg[ width=\widthmultiplierdraw\linewidth]{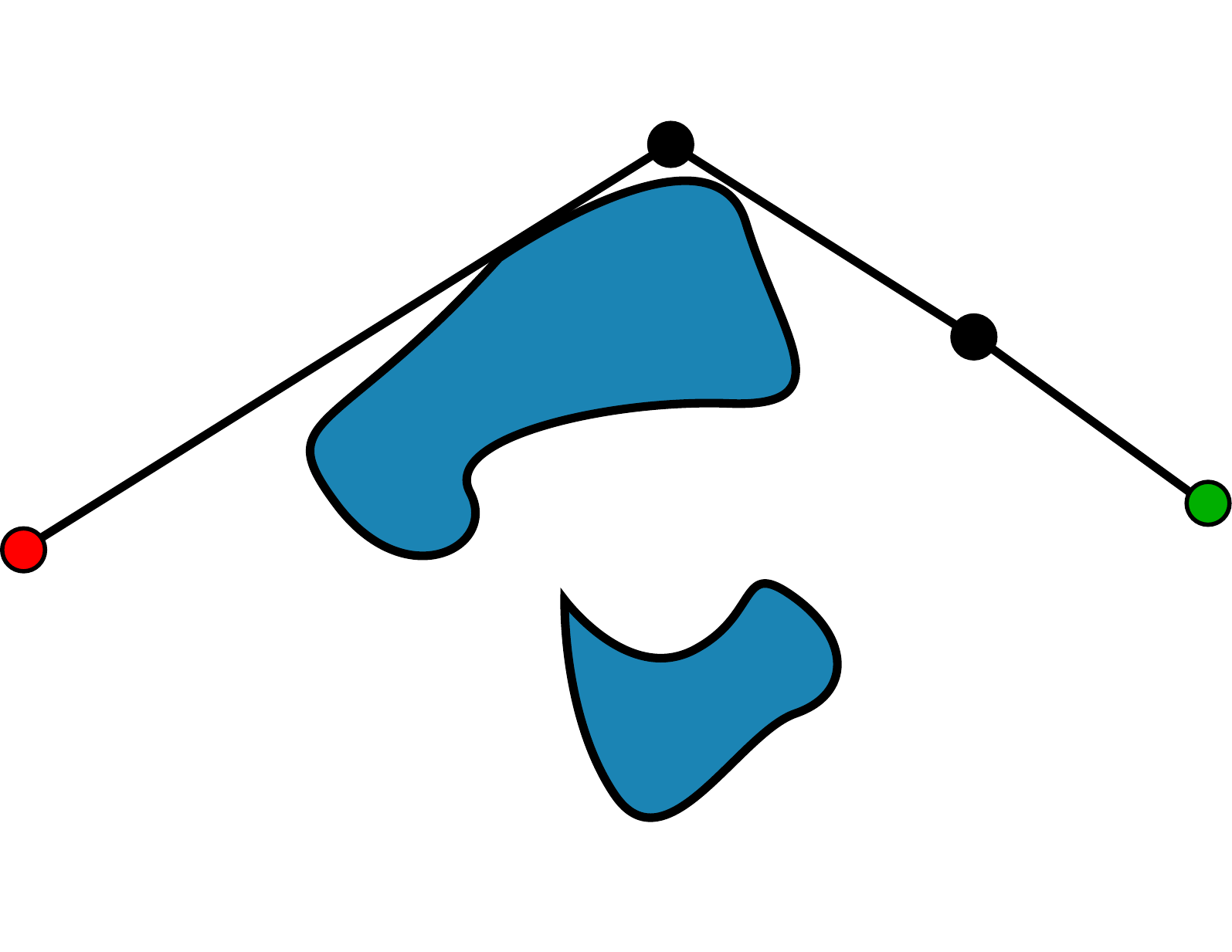}}\label{fig:insert_p}}\hspace{1em}
	\subfloat[Neighbourhood of vertex 2 of $p$ that has been extended to include previous vertex from path.]{\fbox{\includesvg[ width=\widthmultiplierdraw\linewidth]{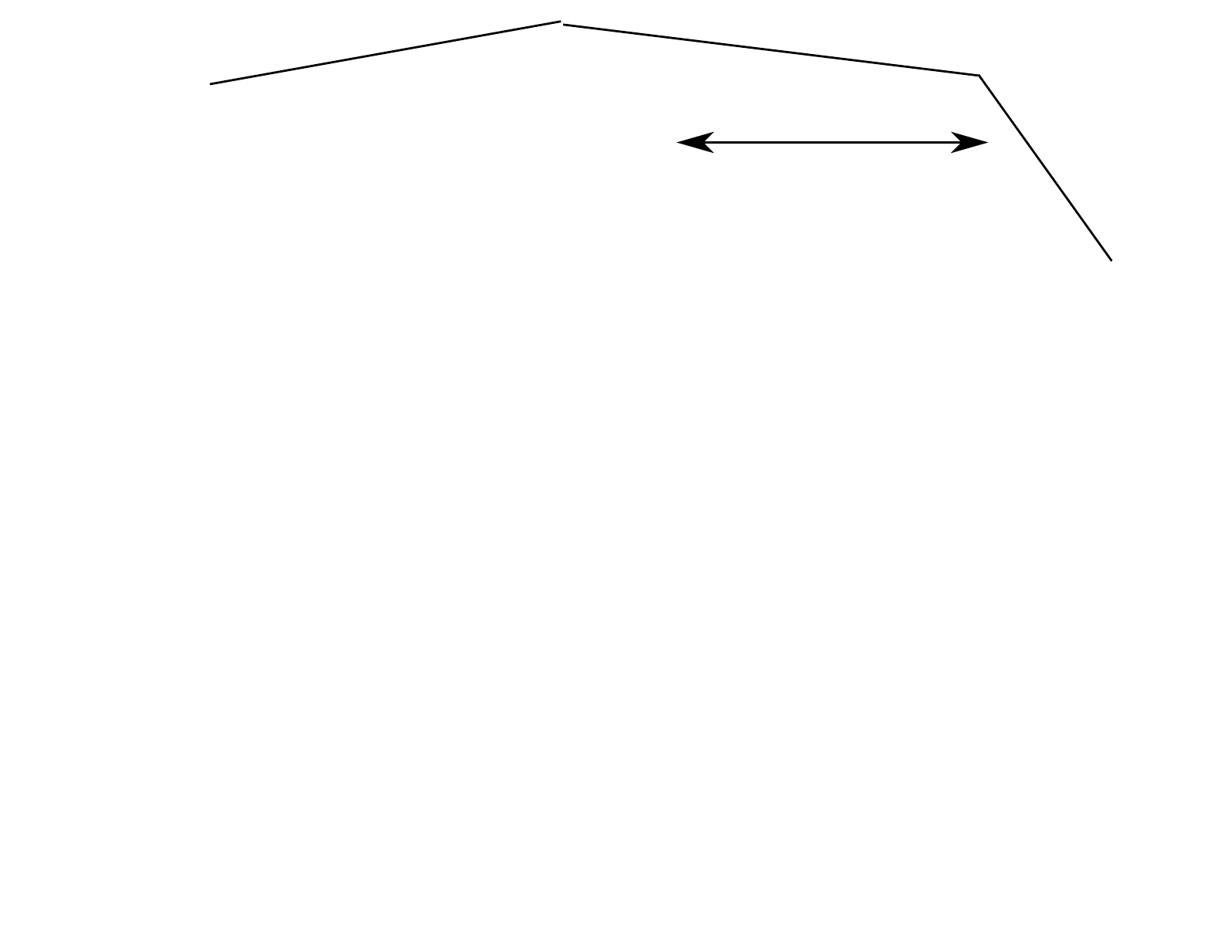}}\label{fig:insert_first}}\hspace{1em}
	\subfloat[The vertex is added to $G$ and its neighbourhood is rewired.]{\fbox{\includesvg[ width=\widthmultiplierdraw\linewidth]{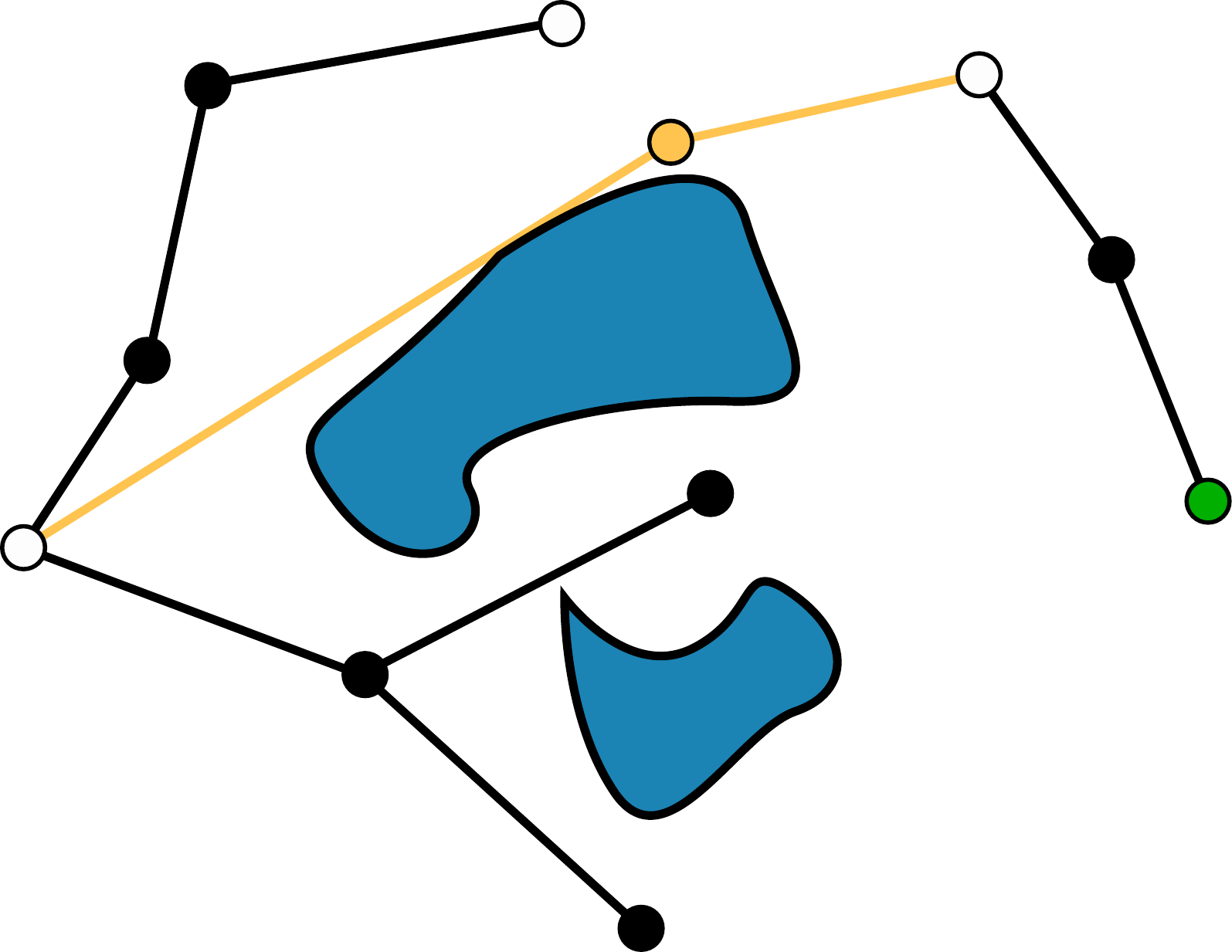}}\label{fig:insert_rewire_first}}
	
	\hspace*{-1.75in}
	\subfloat[Neighbourhood of vertex 3 from $p$ that has been extended to include previous vertex from path.]{\fbox{\includesvg[ width=\widthmultiplierdraw\linewidth]{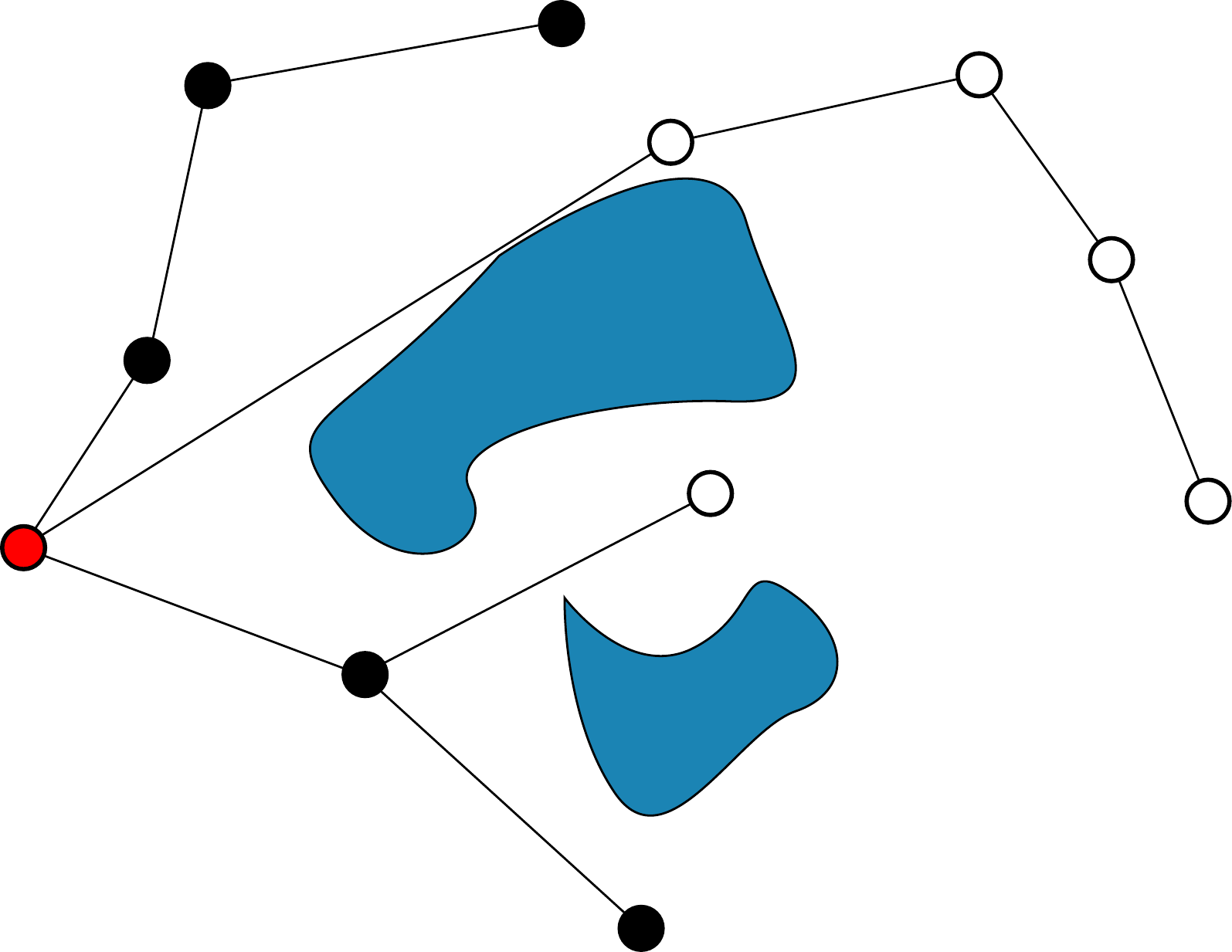}}\label{fig:insert_second}}\hspace{1em}
	\subfloat[The vertex is added and its neighbourhood is rewired.]{\fbox{\includesvg[ width=\widthmultiplierdraw\linewidth]{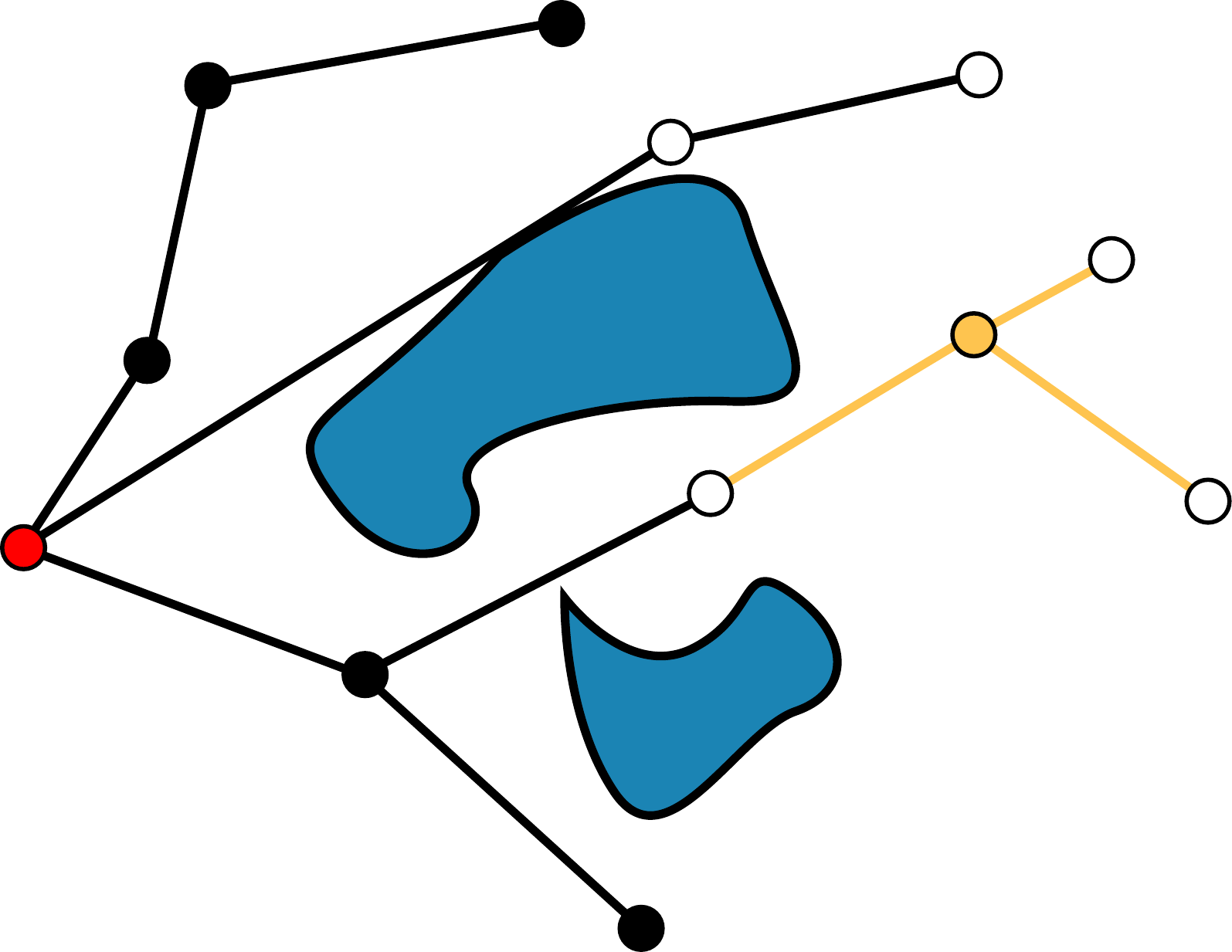}}\label{fig:insert_rewire_second}}\hspace{1em}
	\subfloat[The final vertex of $p$ is inserted into $G$, but no edges change.]{\fbox{\includesvg[ width=\widthmultiplierdraw\linewidth]{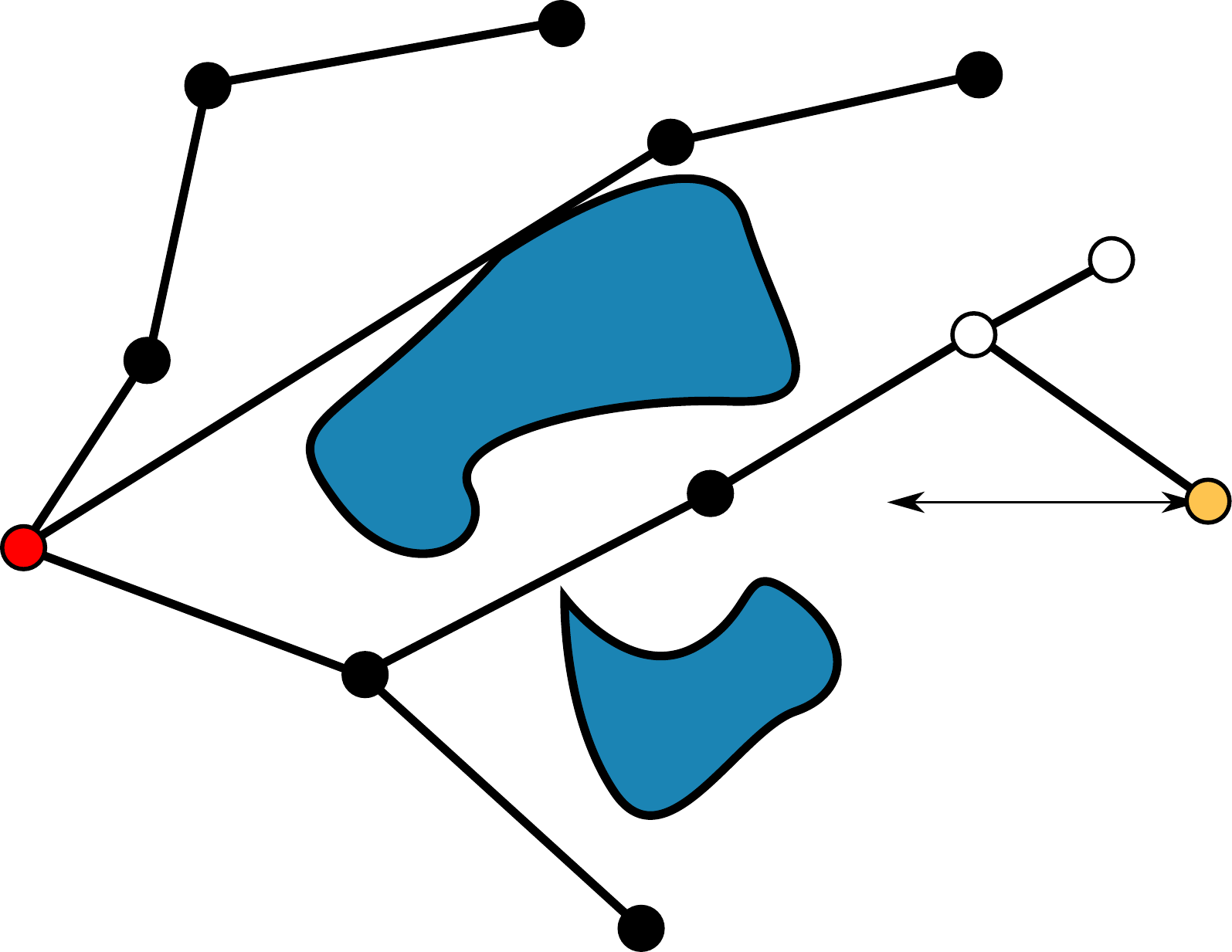}}\label{fig:insert_rewire_third}}\hspace{1em}
	\subfloat[$G$ after $p$ has been added, the new minimum cost path shown in yellow.]{\fbox{\includesvg[ width=\widthmultiplierdraw\linewidth]{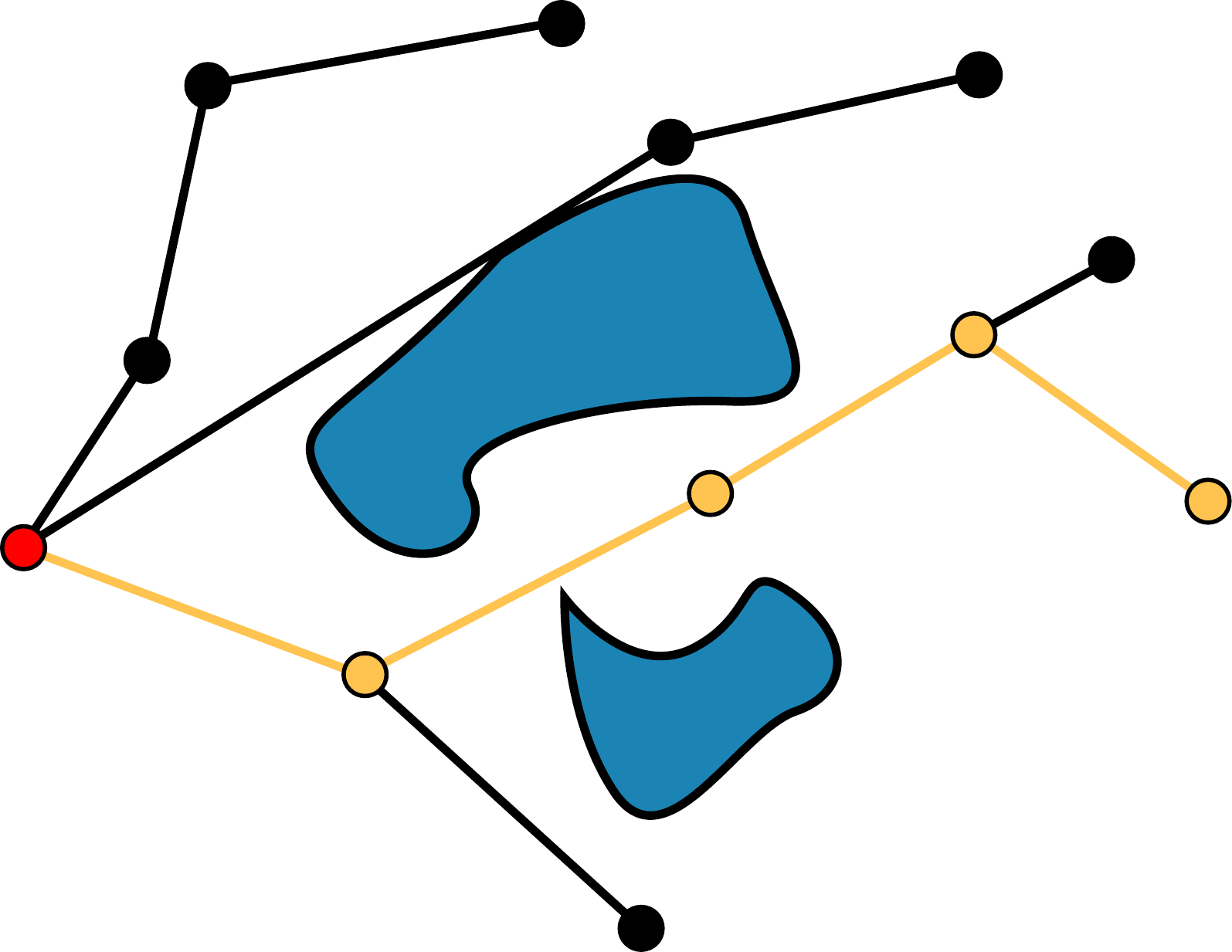}}\label{fig:insert_final_path}}
	\restoregeometry
	\caption{Insertion of a path $p$ into a planner's graph $G$ using RRT*'s insertion procedure for the objective of minimising Euclidean path length. The start vertex in $G$ is red and the goal vertex is green. Vertices and the edges that are added/modified are shown in yellow, except for (h) where the final path is shown in yellow. Vertices part of a yellow vertex's neighbourhood have a white fill. $r\mathsub{n}$ is the radius that defines the neighbourhood of the yellow vertex.}
	
	\label{fig:insert_path_rrtstar}
\end{figure}

To maintain asymptotic optimality, vertices from the short-cut path are rewired into RRTConnect*'s graph. To ensure that the short-cut path is recoverable through RRTConnect*'s graph, the neighbourhood of each of the path's vertices is expanded to include the path's previous vertex as shown in Fig.~\ref{fig:insert_path_rrtstar}. After path insertion the cost of the best path through the planner's graph $C'\mathsub{best}$ is:
\begin{equation}
C'\mathsub{best} \leq \textrm{min}(C\mathsub{best}, C\mathsub{path})
\end{equation}
Where $C\mathsub{path}$ is the cost of the path that was inserted and $C\mathsub{best}$ is the cost of the best cost path before the new path was inserted. In our experiments we terminate the Shortcut routine after a fixed number of iterations.

Our approach can be extended to other planners by changing the planner used in Fig.~\ref{alg:integrated_with_local_opt} (line 4). The InsertPath method may have to be altered for use with planners such as PRM* that do not perform rewiring. 
\FloatBarrier
\section{Experiments}
\begin{figure}[htb]
	\centering
	\subfloat[Vine to be pruned.]{\includegraphics[height=\exptwidth\linewidth]{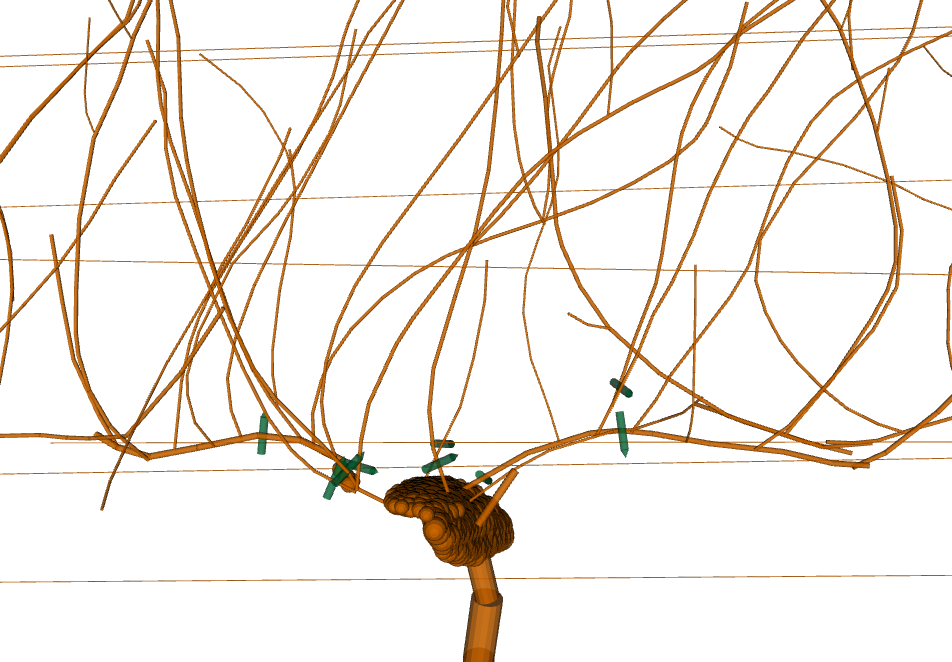}}\hspace{5em}
	\subfloat[The robot arm in a cutting position.]{\includegraphics[height=\exptwidth\linewidth]{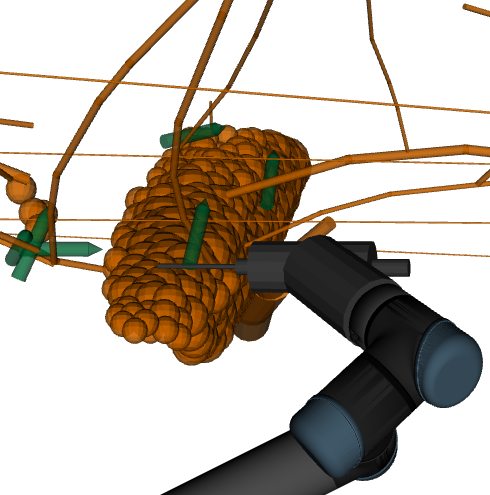}}
	\caption{Vine pruning scenario.}
	\label{fig:vine_pruning_robot}
\end{figure}

\begin{figure}[htb]
	\centering
	\subfloat[Robot arm with gripper model.]{\includegraphics[width=\exptwidth\linewidth]{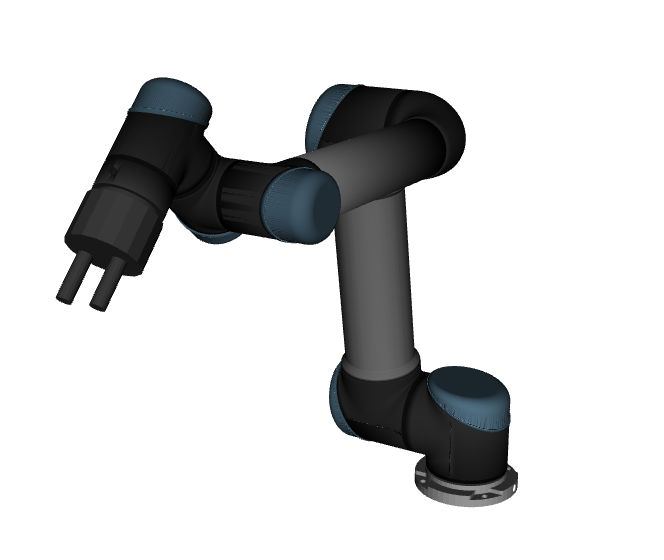}}\hspace{5em}
	\subfloat[Robot arm reaching into a cubicle.]{\includegraphics[width=\exptwidth\linewidth]{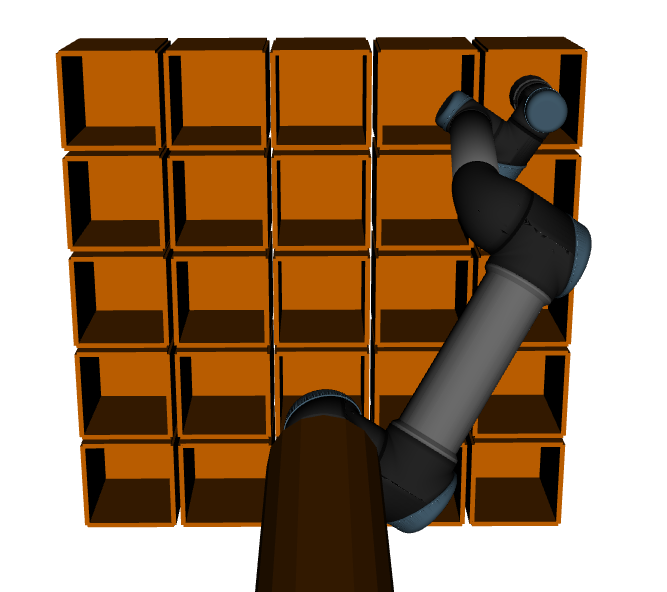}}
	\caption{Cubicle picking scenario.}
	\label{fig:cubicle_picking_robot}
\end{figure}

To test our approach, we compare the performances of the planners in Tab.~\ref{tab:planners} to RRTConnect* integrated with a short-cutting local optimiser. We test these planners on two robots, one for pruning grape vines~\cite{BotterillVPR} (Fig.~\ref{fig:vine_pruning_robot}) and one for reaching into cubicles (Fig.~\ref{fig:cubicle_picking_robot}).

In both trials, RRTConnect* with and without short-cutting was configured to minimise Euclidean path length. Planner parameters are in Appendix~\ref{sec:appendix_parameters} along with how we arrived at these choices. We also tested with RRT* (with and without local optimisation) but it was unable to find initial solutions, which is consistent with previous results when using a robot arm~\cite{Zucker2013}.

On the vine pruning robot, the planners were tasked with moving the robot arm to cut positions on the vine. This task involves planning fine motions around thin obstacles. We used a collision detector that was specialised for use in this problem~\cite{Paulin2015b}.

The cubicle picking environment was designed to be similar to that used in previous research~\cite{POMPChoudhury2016, Phillips2012, Ratliff2009} and the 2015 Amazon Picking Challenge~\cite{Correll2016}. The planner had to compute plans so that the robot arm would reach from its start position in one cubicle into another. Exiting the start cubicle and entering the goal cubicle both required fine motion plans. We used the Flexible Collision Library (FCL)~\cite{Pan2012} for collision detection. An analytical IK solver for the UR5~\cite{Hawkins2013} was used to generate the robot arm configurations to reach the arm into the centre of each cubicle with a fixed end-effector orientation.

\section{Results}
For both experiments we recorded the Euclidean length (sum of Euclidean lengths of each path segment, in radians), execution time (how long it would take the robot arm to follow the path), the number of local optimisations and the cycle time (computation time plus execution time). These measurements were taken from the planner in a separate thread as to not interfere with the planner's performance. Values for length, and execution time of the shortest found path taken before the first solution was found were later calculated using the first solution that the planner found. The total time is the planning time plus execution time, where planning is terminated after $t$ seconds, or once a solution is found if this takes longer.

\begin{figure}
	\centering
	\includegraphics[width=\linewidth]{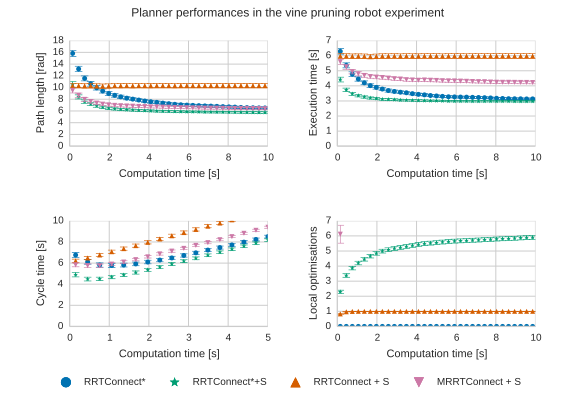}
	\caption{Means for \numsuccessfullcuts successful grape vine planning queries. Error bars show the 95\% confidence interval. MRRTConnect+S averaged 905 local optimisations after 30 seconds of planning, it was truncated for clarity. For a fixed time budget RRTConnect*+S found paths that were faster to execute than RRTConnect*, allowing the robot to have a shorter cycle time.}
	\label{fig:vine_pruning_results}
\end{figure}

\begin{figure}[H]
	\centering
	\includegraphics[width=\linewidth]{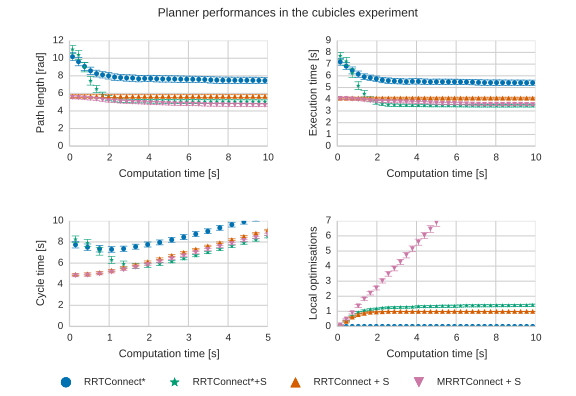}
	\caption{Means for \numcubicles cubicles queries. Error bars show the 95\% confidence interval. MRRTConnect+S averaged 42.3 local optimisations after 30 seconds of planning, it was truncated for clarity. For a fixed time budget RRTConnect*+S found paths that were faster to execute than RRTConnect*, allowing the robot to have a shorter cycle time.}
	\label{fig:cubicle_picking_results}
\end{figure}

Integrating RRTConnect* with a short-cut local optimiser resulted in significant speed-ups as shown in Fig.~\ref{fig:vine_pruning_results} and Fig.~\ref{fig:cubicle_picking_results}. It resulted in a 24\% reduction in cycle time for the vine pruning robot, and a 21\% decrease in cycle time for the cubicle picking robot. MRRTConnect+S also performed well in both experiments. These speed-ups result in lower robot cycle times because the computation and execution times are of similar magnitude.

\FloatBarrier
\section{Discussion}
Integrating RRTConnect* with a short-cut local optimiser resulted in shorter paths being found more quickly compared to not using the short-cut optimiser as shown in Fig.~\ref{fig:vine_pruning_results} and Fig.~\ref{fig:cubicle_picking_results}. This is consistent with the results of a recent preprint~\cite{Kuntz2016} where BIT* and PRM* were interleaved with a Lagrangian local optimiser. 

In the cubicles experiment the RRTConnect*+S planner only performed around one local optimisation. This is because it tended to find short solutions after one local optimisation and could not improve these solutions enough to invoke the local optimiser again. RRTConnect+S also performed well on this experiment. This suggests that the configuration space for the cubicles experiment is very sparse and optimising a wide range of initial paths could result in a short path. 

RRTConnect*+S was sparing with its use of the local optimiser in both experiments as shown in Fig.~\ref{fig:vine_pruning_results} and Fig.~\ref{fig:cubicle_picking_results}. This is because the local optimiser is only invoked when RRTConnect* has improved the path by a certain threshold (Fig.~\ref{alg:integrated_with_local_opt}). MRRTConnect+S made a lot of calls to the local optimiser because it was called every time a new path was found. We might also expect the interleaving optimiser~\cite{Kuntz2016} to make a lot of calls to the local optimiser because it is invoked every time the global planner (even slightly) improves the path. MRRTConnect+S and the interleaving approach may spend a lot of time in local optimisation if a slow local optimiser is used.

The short-cut optimiser was a good fit for both the experiments as shown by the good performance of MRRTConnect+S in both experiments. This could be caused by the robots having sparse configuration spaces in both experiments. The short-cut optimiser is not a good fit for all problems, especially those where the triangle inequality does not hold. In these spaces it is possible that short-cutting a path may lead to it becoming longer. Our path insertion method (see Sec.~\ref{sec:integration}) guarantees that the insertion of a poor path does not degrade the quality of any other paths found by the planner. 

\section{Conclusion}
We presented an approach to integrating an asymptotically optimal path planner with a local optimiser. In our experiments we saw that integrating a short-cutting local optimiser significantly improved the performance of RRTConnect* in two robot arm tasks. Our approach resulted in a significant performance improvement when compared with the state-of-the-art RRTConnect* asymptotically optimal planner and computes paths that are 31\% faster to execute when both are given 3 seconds of planning time.

\bibliography{library}

\section{Appendix A Parameters used}
\label{sec:appendix_parameters}
In both experiments the parameters were selected after performing a parameter sweep on a subset of planning queries that were not used in the test set. The parameters used in each experiment are listed in Tab.~\ref{tab:vine_pruning_robot_parameters_used} and Tab.~\ref{tab:cubicle_reaching_parameters}. In both cases RRTConnect* was set to use informed sampling and it rejected new states that could not improve the cost of the current graph (i.e. focus search enabled in OMPL). Range is a parameter of RRTConnect*. When the short-cut local optimiser was invoked it was set to perform $i$ iterations:
\begin{equation}
i = \textrm{SCF} * N\mathsub{vertices}
\end{equation}
Where SCF is the Shortcut count factor parameter and $N\mathsub{vertices}$ is the number of vertices in the path. The local optimisation threshold is a parameter of our proposed approach and appears in Fig.~\ref{alg:integrated_with_local_opt}.

\begin{table}[!h]
	\centering
	\caption{Parameters used in vine pruning robot tests}
	\begin{tabular}{l | l | l | l}
		Planner & range & shortcut count factor & local optimization threshold \\ \hline
		RRTConnect* & 2.5 & - & - \\
		RRTConnect* with shortcut & 2.5 & 3.0 & 0.01 \\
		RRTConnect+S & 0.5 & 4.0 & - \\
		MRRTConnect+S & 0.5 & 4.0 & - \\
	\end{tabular}
	\label{tab:vine_pruning_robot_parameters_used}
\end{table}

\begin{table}[!h]
	\centering
	\caption{Parameters used in cubicle picking tests}
	\begin{tabular}{l | l | l | l}
		Planner & range & shortcut count factor & local optimization threshold \\ \hline
		RRTConnect* & 0.5 & - & - \\
		RRTConnect* with shortcut & 3.0 & 3.0 & 0.11 \\
		RRTConnect+S & 0.5 & 3.0 & - \\
		MRRTConnect+S & 0.5 & 3.0 & - \\
	\end{tabular}
	\label{tab:cubicle_reaching_parameters}
\end{table}

\end{document}